\documentclass[10pt,twocolumn,letterpaper]{article}

\usepackage{wacv}
\usepackage{times}
\usepackage{epsfig}
\usepackage{graphicx}
\usepackage{amsmath}
\usepackage{amssymb}
\usepackage{bm}
\usepackage{fixltx2e}
\usepackage{placeins}
\usepackage{gensymb}
\usepackage{tablefootnote}
\usepackage[font=footnotesize]{caption}
\usepackage[pagebackref=true,breaklinks=true,letterpaper=true,colorlinks,bookmarks=false]{hyperref}
\wacvfinalcopy % *** Uncomment this line for the final submission

\def\wacvPaperID{****} % *** Enter the wacv Paper ID here

\ifwacvfinal\fi
\begin{document}

%%%%%%%%% TITLE
\title{Fast Face Image Synthesis with Minimal Training}

% Authors at the same institution
%\author{First Author \hspace{2cm} Second Author \\
% Authors at different institutions

\author{\parbox{16cm}{\centering
    {\large Sandipan Banerjee, Walter J. Scheirer, Kevin W. Bowyer, and Patrick J. Flynn}\\
    Department of Computer Science \& Engineering, University of Notre Dame, USA\\
        \tt\small \{sbanerj1, wscheire, kwb, flynn\}@nd.edu
    }
    % <-this % stops a space
}

% \author{Sandipan Banerjee \\
% University of Notre Dame, IN\\
% {\tt\small sbanerj1@nd.edu}
% \and
% Walter J. Scheirer \\
% University of Notre Dame, IN\\
% {\tt\small walter.scheirer@nd.edu}
% \and
% Kevin. W. Bowyer \\
% University of Notre Dame, IN\\
% {\tt\small kwb@nd.edu}
% \and
% Patrick. J. Flynn \\
% University of Notre Dame, IN\\
% {\tt\small flynn@nd.edu}
% }

\maketitle
\ifwacvfinal\thispagestyle{empty}\fi

%%%%%%%%% ABSTRACT
\begin{abstract}
We propose an algorithm to generate realistic face images of both real and synthetic identities (people who do not exist) with different facial yaw, shape and resolution. The synthesized images can be used to augment datasets to train CNNs or as massive distractor sets for biometric verification experiments without any privacy concerns. Additionally, law enforcement can make use of this technique to train forensic experts to recognize faces. Our method samples face components from a pool of multiple face images of real identities to generate the synthetic texture. Then, a real 3D head model compatible to the generated texture is used to render it under different facial yaw transformations. We perform multiple quantitative experiments to assess the effectiveness of our synthesis procedure in CNN training and its potential use to generate distractor face images. Additionally, we compare our method with popular GAN models in terms of visual quality and execution time.
\end{abstract}

\section{Introduction}\label{sec:Intro}
Face image synthesis has been a popular research area recently \cite{SynthesisSurvey}, mainly as a means to generate artificial training samples for CNNs \cite{DLNature}. Generative adversarial nets (GANs) \cite{GAN} have made tremendous progress in this domain with different GAN models being used to generate synthetic face images 
with different pose \cite{DRGAN,CVAEGAN,NIPS17}, facial feature \cite{DCGAN,BEGAN,MAGAN,EBGAN}, age \cite{cGAN,FaceAge} and expression \cite{coGAN}. However, as pointed out by Karras \etal \cite{ProgressiveGAN}, GANs require plenty of training data (10M images), time (19 days) and GPU resources to generate high quality (1024$\times$1024) face images. We propose an algorithm that requires few gallery images (a few thousand), minimal training (of an SVM \cite{SVM}) and little execution time (a few seconds) to generate high quality synthetic face images with different pose and shape of real or synthetic identities. 

\begin{figure}[t]
\centering
   \includegraphics[width=1.0\linewidth]{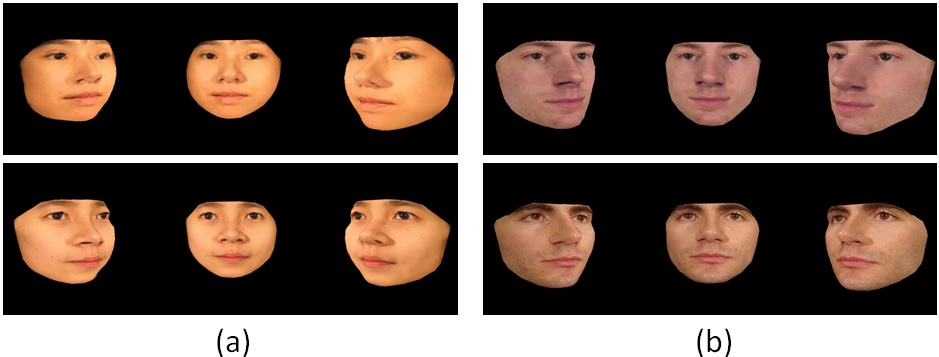}
   \caption{Sample face images (512$\times$512) for - (a) Female Asian, and (b) Male Caucasian ethnic groups. Each row represents three face images of the same identity at yaw values -30\degree, 0\degree and +30\degree. The face images in each top row belong to a synthetic identity while the corresponding bottom row has face images of a real identity for visual comparison.}
\label{fig:Teaser1}
%\label{fig:onecol}
\end{figure}

Our method begins by pooling real, frontal gallery face images similar in appearance and combines facial components from these gallery images to generate realistic synthetic textures. Compatible 3D face shape models, obtained by performing a best-fitting test, are then used to render the synthetic texture in 3D with different pose and shape, where shape is a function of facial landmarks \cite{MasiAug}. Our methodology remains the same even for different target resolutions (100$\times$100 to 800$\times$600). Some of the faces in Figure \ref{fig:Teaser1} belong to synthetic identities generated using our method.

% \footnotetext{Synthetic identities in Figure \ref{fig:Teaser1} - (a) bottom row, (b) top row, (c) top row, (d) bottom row.}

The synthetic images generated by our method can be utilized in the following ways:\\ 
(1) Researchers have shown that an increase in the number of training images per identity can boost the representation capability of CNNs \cite{MasiAug}. Also, for two datasets with the same number of face images, the one with more identities offers more diversity to a CNN to learn representations from, and hence achieves a higher performance \cite{DosDonts}. Therefore, training datasets need to capture plenty of intra- and inter-subject variance to perform well. Our method can generate new images of an existing identity (new texture, pose and shape) and synthetic identities to boost the intra- and inter-subject variation.\\
(2) As pointed out by \cite{MegaFace,MF2}, the recognition performance drops for a CNN when a large number of distractor face images is introduced into the gallery. Our method can potentially create a synthetic distractor dataset, of any desired size, that can be used without running into potential issues of identity overlap or invasion of privacy. A dataset created with such synthetic face images will be free from any labeling errors like those that are found in public datasets such as VGG-Face \cite{DosDonts,UMDFaces}. \\
(3) The synthetic face images can be used by law enforcement to train forensic experts to recognize faces from a diverse set of individuals.

We perform three quantitative benchmark experiments with our synthetic face images - i) a verification experiment testing on the IJB-B dataset \cite{IJBB} using the VGG-16 network \cite{Simonyan14c} fine-tuned on hybrid datasets consisting of both real (from \cite{CASIA}) and synthetic face images (generated by our method), ii) a validation experiment with VGG-Face \cite{VGG} and ResNet-101 \cite{ResNet} networks using real and synthetic distractor sets, testing on the FG-Net dataset \cite{FGNet}, and iii) a timing and visual quality comparison with three popular GAN models (DCGAN \cite{DCGAN}, BEGAN \cite{BEGAN}, PGGAN \cite{ProgressiveGAN}) while generating 128$\times$128 synthetic face images. Results show our synthetic face images to be stable and effective as supplements to existing datasets, and can be used as distractors in face recognition experiments. 

\section{Related Work}
{\bf Face recognition:} The rise of deep learning \cite{DLNature} has seen tremendous improvements in face recognition performance on previously challenging datasets like LFW \cite{LFW}, with the focus now shifted towards video based face recognition \cite{IJBA,IJBB,PaSC,YoutubeFaces}. Massive face image datasets, both private \cite{Facebook_Deepface,Google_FaceNet} and public \cite{VGG,VGGFace2,UMDFaces}, have been accumulated to train intricately developed CNN models \cite{ResNet,LightCNN,SENet}. Researchers have used multi-pose based CNN models to recognize face images with multiple yaw values during testing \cite{AbdAlmageed2016multipose,masi2016cvpr} or normalized facial pose in an attempt to stabilize the data \cite{TMM15,TIST16} or used artificial pose synthesis as a means for training data augmentation \cite{MasiAug}. The ultimate goal is to construct a CNN model which is robust to massive sets of distractor face images \cite{MegaFace,MF2}.

{\bf Face Image Synthesis:} Research in this domain began by recombining neighborhood patches to hallucinate new faces \cite{ICME05}. More structured methods have been formulated since then, like stitching similar faces or their parts together \cite{Bitouk08,ACCV14,SREFI1}, or swapping face masks \cite{FaceSwap,IraPortrait} and expressions \cite{SIGGRAPH09,SIGGRAPH11} onto different background images for face image synthesis and inpainting. The use of a 3D head model to repose or frontalize a given face image in order to generate synthetic views and their impact on face recognition have been investigated in \cite{MasiAug,HeFrontal,LiuFrontal,USC3DMM,Bulat3D,HassFront,Vito}. While GANs have made their mark on face synthesis research, other deep feature based techniques have been explored by researchers for reconstructing a face image \cite{Belanger,DFI,PixelNN,FaceSketch,SketchBMVC17}. A more detailed bibliography can be found in \cite{SynthesisSurvey}. 
\section{The Proposed Approach}
On a high level, our face image synthesis method starts with a real frontal face image (the base face), whose facial region is triangulated using landmark points. We replace the triangles of this face image with corresponding triangles from other real face images or donor images, similar in appearance, and blend the triangles together to obtain a synthetic face texture and filter by visual quality. To render this texture in 3D, at different yaw values, we use best-fitting 3D models using texture and shape parameters. 

\subsection{Synthetic Texture Generation}\label{sec:TexFull}
In this section we describe our texture generation pipeline. We use a subset of the public dataset in \cite{SREFIDonor}, comprised of 15,807 face images of 1,352 identities, to synthesize these textures. The distribution of this subset can be seen in Table \ref{Tab:SREFIData}. All the images were aligned about their eye centers and resized to 512$\times$512 beforehand. Our texture synthesis pipeline takes one to four seconds to execute on average, depending on the target resolution.\\

\noindent {\bf Finding the Number of Donors Necessary for Anonymization.} The method in \cite{SREFI1} replaces facial regions with seven to ten donors to create a synthetic texture based on gallery size. However, this \emph{ad hoc} technique may lead to natural looking texture when the gallery is large; for a smaller gallery, the generated texture can look distinctly non-uniform as we transition from one larger face part to another. Therefore, we regularize the donor selection process by determining the best number of donors, {\bf N}.

\begin{figure*}
\centering
   \includegraphics[width=1.0\linewidth]{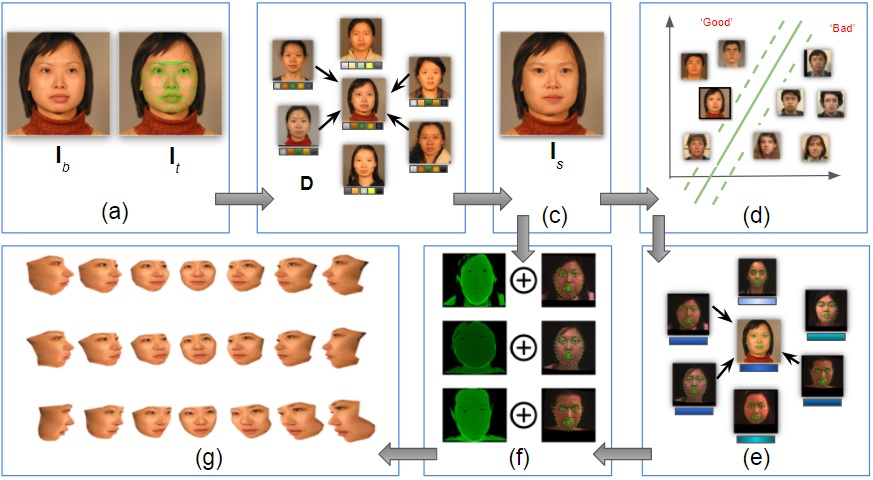}
   \caption{Our face image synthesis pipeline - (a) The input base face $\text{{\bf I}}_{b}$ is triangulated using landmark points, (b) its donor pool {\bf D} is constructed using hypercolumn feature maps, represented by color bars, (c) the synthetic texture $\text{{\bf I}}_{s}$ is created using images from {\bf D}, (d) $\text{{\bf I}}_{s}$ is subjected to quality based filtering, (e) best fitting 3D models are obtained using proximity in landmark and deep feature (represented by blue bars), (f) dense triangular mesh is generated from corresponding 3D models, (g) multi-pose and shape renderings of $\text{{\bf I}}_{s}$.}
\label{fig:Pipeline}
%\label{fig:onecol}
\end{figure*}

We choose a random set of 1,545 face images of 100 identities from our gallery \cite{SREFIDonor} to gauge the optimal {\bf N}. The facial region of each image was triangulated using Delaunay's method from landmark points extracted using Dlib \cite{Dlib}. We shifted these triangles using their centroids to separate facial regions like the eyes, nose, mouth from each other \cite{SREFI1}. We replace triangles with one donor assigned per region, to anonymize the original face image \emph{i.e.} change its identity. We vary donor size {\bf N} from 0 (original image) to 5 (swapping regions with 5 donors) and use normalized \emph{fc7} layer features from the pre-trained VGG-Face network \cite{VGG} to match the set of synthesized faces with the original set (using cosine similarity). We use the True Accept Rate (TAR) at False Accept Rate (FAR) of 0.01 as our performance metric. For {\bf N} = 0, the face images match each other very well with TAR over 0.96 but at {\bf N} = 4 the TAR drops precipitously to only 0.12. Therefore, replacement with 4 donors can anonymize the original face image and result in a new synthetic identity.\\

\noindent {\bf Pooling Proximal Faces: Hypercolumns.} Given a base face image, we construct its donor pool {\bf D} with face images of potential donor identities. We choose these identities based on proximity in feature space from the base face. For a gallery face image we extract its hypercolumn descriptor using \emph{conv}-[$1_2$,$2_2$,$3_3$,$4_3$,$5_3$] feature maps from pre-trained VGG-Face \cite{VGG}. We use hypercolumn features as they capture information at different spatial contexts instead of the high level features of the \emph{fc7} layer \cite{SREFI1}. However, the hypercolumn feature maps extracted consist of 434 dimensions for each pixel of an image. Hence, for a 512$\times$512 gallery image {\bf I}, a 512$\times$512$\times$434 hypercolumn vector $\text{{\bf V}}_{I}$ is obtained. To reduce computation time and feature redundancy, we sample a 68$\times$434 feature map of $\text{{\bf V}}_{I}$ from the 68 landmark points (pixels) of the face, obtained using Dlib \cite{Dlib}. A mean vector $\text{{\bf S}}_{I}$ for each identity is obtained by computing the average of $\text{{\bf V}}_{I}$ for all its images. We calculate the distance between any two identities as follows:

\begin{equation}
d(I1,I2) = \sum_{i=1}^{68}\sum_{j=1}^{434}\left | (\text{{\bf S}}_{I1})_{ij} - (\text{{\bf S}}_{I2})_{ij}\right |,
\label{eq:pooling}
\end{equation}

\noindent where $I1$ and $I2$ are two gallery identities, $\text{{\bf S}}_{I1}$ and $\text{{\bf S}}_{I2}$ are their mean sampled hypercolumn feature maps and $d(I1,I2)$ is the distance between them. For the base face, we cluster identities with the lowest $d$ values and construct its donor pool {\bf D} with their face images. If creating synthetic texture of a real identity, {\bf D} is simply composed of all images of that identity in the gallery. For each gallery identity, {\bf D} is constructed and stored as an offline step. \\

\begin{table*}
\caption{Data Distribution of the Gallery and Synthetic Datasets}
\begin{center}
\captionsetup{justification=centering}
\begin{small}
\begin{tabular}{| l | c | c | c | c | c | c |}
\cline{2-7}
\multicolumn{1}{c|}{}& \multicolumn{2}{c|}{\begin{tabular}[x]{@{}c@{}}{\bf Gallery Images \cite{SREFIDonor}}\\(identities)\end{tabular}} & \multicolumn{2}{c|}{\begin{tabular}[x]{@{}c@{}}{\bf Synthetic Textures}\\(identities)\end{tabular}} & \multicolumn{2}{c|}{\begin{tabular}[x]{@{}c@{}}{\bf Synthetic 3D Images}\\ (identities)\end{tabular}}\\
\hline
{Ethnicity} & {Male} & {Female} & {Male} & {Female} & {Male} & {Female} \\
\hline
Caucasian & {\begin{tabular}[x]{@{}c@{}}7,108\\(678)\end{tabular}} & {\begin{tabular}[x]{@{}c@{}}5,510\\(600)\end{tabular}} & {\begin{tabular}[x]{@{}c@{}}47,168\\(6,398)\end{tabular}} & {\begin{tabular}[x]{@{}c@{}}35,127\\(4,527)\end{tabular}} & {\begin{tabular}[x]{@{}c@{}}972,511\\(6,398)\end{tabular}} & {\begin{tabular}[x]{@{}c@{}}730,540\\(4,527)\end{tabular}}\\
\hline
Asian & {\begin{tabular}[x]{@{}c@{}}1,903\\(100)\end{tabular}} & {\begin{tabular}[x]{@{}c@{}}1,286\\(74)\end{tabular}} & {\begin{tabular}[x]{@{}c@{}}9,206\\(820)\end{tabular}} & {\begin{tabular}[x]{@{}c@{}}8,261\\(593)\end{tabular}} & {\begin{tabular}[x]{@{}c@{}}188,116\\(820)\end{tabular}} & {\begin{tabular}[x]{@{}c@{}}169,825\\(593)\end{tabular}}\\
\hline
\end{tabular}
\label{Tab:SREFIData}
\end{small}
\end{center}
\end{table*}

\noindent {\bf Stitching Triangles Together.} Once the donor pool {\bf D} is constructed, we randomly select four donors to replace triangulated regions of the base face with a specific donor only assigned to a particular region. Prior to replacement, each donor triangle is randomly reshaped with parameters within the inter-quartile range of permissible shapes biologically possible for a given gender and race \cite{SREFI1}. Each triangle is then color adjusted by shifting the mean RGB values of its pixels to be same as that of the base face triangle it replaces. The donor triangle is then overlaid on the base face and blended in using Laplacian pyramids \cite{LapPyr}. This approach blends pixels from the donor image $\text{{\bf I}}_{d}$ for the triangle mask {\bf M} at different resolutions of the base face $\text{{\bf I}}_{b}$ situated at different levels of the Laplacian pyramid as:

\begin{equation}
    \text{{\bf I}}_{s} = \text{{\bf M}}\circ \text{{\bf I}}_{b} + (1-\text{{\bf M}})\circ \text{{\bf I}}_{d},
\label{eq:blending}
\end{equation}

\noindent where $\circ$ denotes element wise product and $\text{{\bf I}}_{s}$ is the blended image. The output synthetic face image is generated by collapsing the pyramid top-down and blending all triangles one by one from all four donors.\\   

\noindent {\bf Quality Estimation.} To discard unnatural looking synthetic textures, we implemented an SVM \cite{SVM} based filter. We generated an initial batch of synthetic textures which were rated as `good' or `bad' by non-expert human raters based on visual quality. The textures rated bad generally either had an unnatural shape of facial parts or artifacts created by landmarking errors. A simple binary-class SVM (linear kernel, c = 0.8) was then trained with deep features (\emph{fc7} layer of pre-trained VGG-Face \cite{VGG}) of 2,059 such rated synthetic images, 1,319 of which were marked `bad' and the rest as `good'. The SVM was intentionally biased towards the `bad' side to reduce false negatives. Since generating a new texture is computationally cheap, the SVM can afford to mis-classify some `good' images. This trained SVM is then used as the filtering mechanism in our pipeline. We generate 99,762 textures (frontal face images) for 12,338 synthetic identities from the 15,807 gallery images, as shown in Table \ref{Tab:SREFIData}.

\begin{figure*}
\centering
   \includegraphics[width=1.0\linewidth]{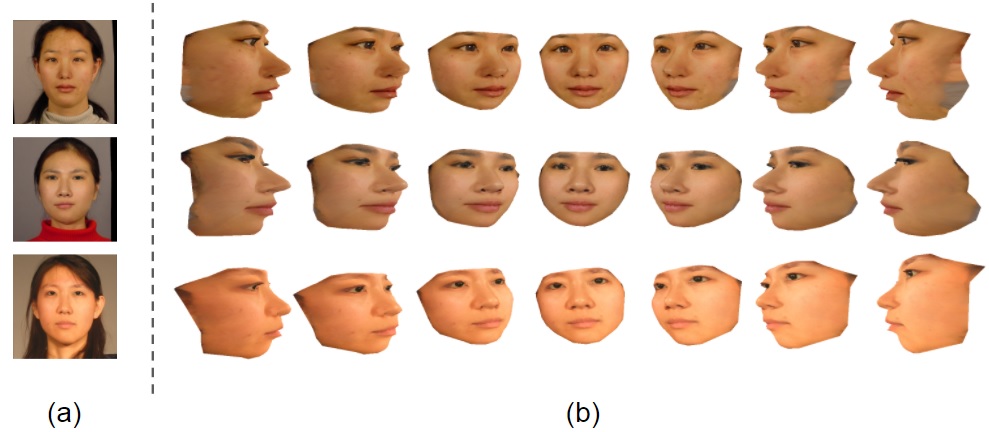}
   \caption{Sample face images generated using our method for three synthetic Female-Asian subjects. The 512$\times$512 synthetic face textures (2D) are shown in the leftmost column (a), with the corresponding 800$\times$600 3D renderings presented to its right (b). The artifact in the top row at facial yaw of -90$\degree$ is due to faulty landmarking.}
\label{fig:SuppFA}
%\label{fig:onecol}
\end{figure*}
\begin{figure*}
\centering
   \includegraphics[width=1.0\linewidth]{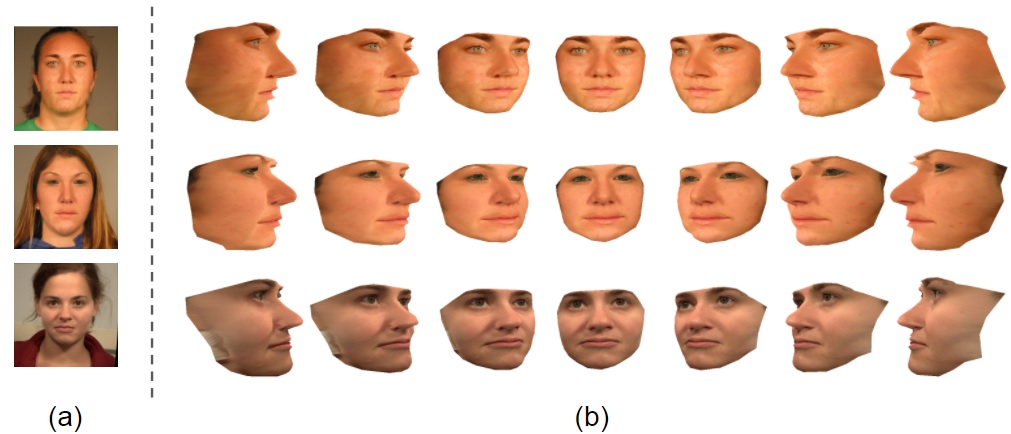}
   \caption{Sample face images generated using our method for three synthetic Female-Caucasian subjects.}
\label{fig:SuppFW}
%\label{fig:onecol}
\end{figure*}
\begin{figure*}
\centering
   \includegraphics[width=1.0\linewidth]{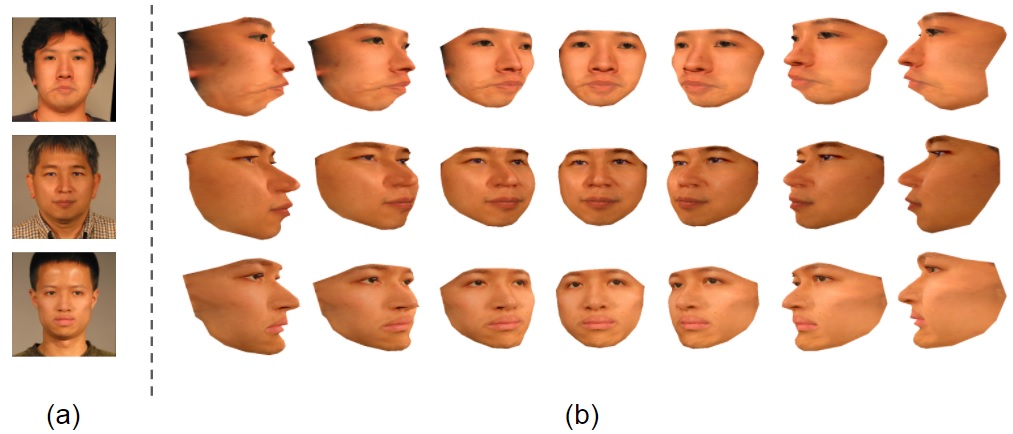}
   \caption{Sample face images generated using our method for three synthetic Male-Asian subjects.}
\label{fig:SuppMA}
%\label{fig:onecol}
\end{figure*}
\begin{figure*}
\centering
   \includegraphics[width=1.0\linewidth]{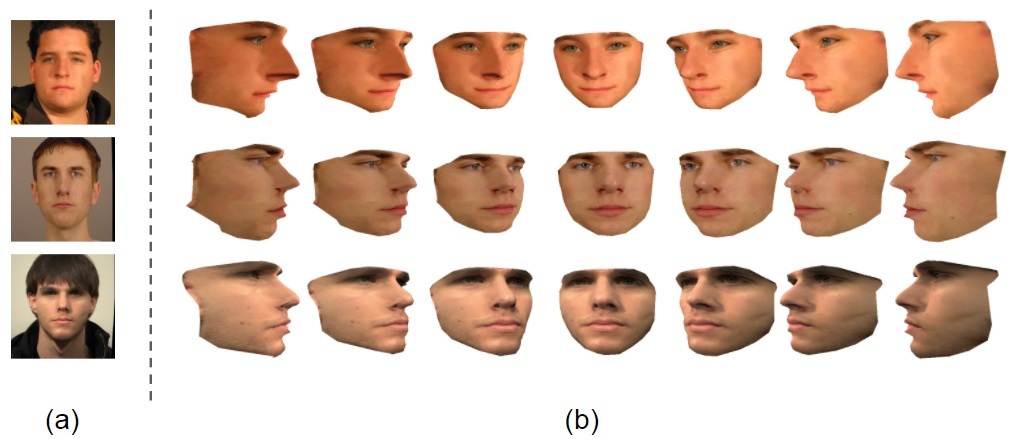}
   \caption{Sample face images generated using our method for three synthetic Male-Caucasian subjects.}
\label{fig:SuppMW}
%\label{fig:onecol}
\end{figure*}

\subsection{3D Face Mask Construction}\label{sec:3DFull}
Here we describe our 3D rendering pipeline to produce multi-pose views of the synthetic textures generated (in Section \ref{sec:TexFull}). Instead of using a generic 3D model \cite{HassFront,MasiAug} or a 3D morphable model (3DMM) \cite{USC3DMM}, we find the `best' fitting 3D model for a synthetic texture from a large set of 3D models to render the texture in 3D. This produces higher quality and more distinct visual results compared to using a generic 3D model and is much faster computationally (two to four secs to generate a face mask), than 3DMM based rendering.\\

\noindent {\bf 3D Models.} We use a set of 3D head images acquired using a Konica-Minolta `Vivid 910' 3D scanner \cite{FaltemierCVIU,SREFIDonor} as models for face shape. The set consists of over 14,000 different 3D face images (640x480 point clouds) accompanied by their corresponding registered color images (2D scans). A majority of the scans are of a near-frontal face with neutral expression (mouth closed). Since our texture synthesis pipeline generates frontal face images with neutral countenance, we discard scans with non-frontal pose or an open mouth. We extract the yaw ($\theta$) of each 2D scan using the CNN model from \cite{PACNN} and remove scans with $\|$$\theta$$\|$ $>$ 10\degree. A two-pass filtering, first using the pre-trained AFFACT network \cite{Affact} and then manual inspection, is implemented to remove scans with the subject's mouth open. Consequently, we end up with 8,462 near-frontal, neutral 3D face models, grouped by gender (Male/Female) and race (Asian/Caucasian) using the metadata available.\\

\noindent {\bf Finding the `Best' Fitting 3D Model.} To find the best-fitting 3D model for a synthetic texture $\text{{\bf I}}_{s}$, we first align each 2D scan (with same race and gender as $\text{{\bf I}}_{s}$) using Dlib \cite{Dlib}. Then the aligned scans are resized to 512x512, same as $\text{{\bf I}}_{s}$. We extract a vector $\text{{\bf V}}_{l}$ for $\text{{\bf I}}_{s}$ and each 2D scan comprising of its 68 landmark points $p \in \mathbb{R}^{2}$ detected using Dlib \cite{Dlib}. We also extract the 4096 dimensional feature vector $\text{{\bf V}}_{f}$ by feeding $\text{{\bf I}}_{s}$ and each 2D scan to the pre-trained VGG-Face network \cite{VGG}. Hence, $\text{{\bf V}}_{l}$ and $\text{{\bf V}}_{f}$ together quantify the overall shape and textural appearance of each face image. To find the best fitting 3D models for $\text{{\bf I}}_{s}$ we use the following equation:

\begin{equation}
dist(\text{{\bf I}}_{s},B) = w_{1} * \left | \text{{\bf V}}_{l}(\text{{\bf I}}_{s}) - \text{{\bf V}}_{l}(B) \right | + w_{2} * \left | \text{{\bf V}}_{f}(\text{{\bf I}}_{s}) - \text{{\bf V}}_{f}(B) \right |,
\label{eq:best3D}
\end{equation}

\noindent where $B$ is a 2D scan of the same gender-race as $\text{{\bf I}}_{s}$ and $dist(\text{{\bf I}}_{s},B)$ captures the dissimilarity in their visual appearance \cite{Belanger}. Since $B$ is of the same gender and race as $\text{{\bf I}}_{s}$, we set $w_{1}$ = 10 and $w_{2}$ = 1 to focus more on facial shape than textural appearance. We assert the 3D model corresponding to the 2D scan ($B$) which minimizes $dist(\text{{\bf I}}_{s},B)$ as the best fitting model for $\text{{\bf I}}_{s}$.\\

\noindent {\bf Rendering in 3D.} We render $\text{{\bf I}}_{s}$ with its three best-fitting 3D models \emph{i.e.} which produce the three minimal $dist$ values, using OpenGL \cite{OpenGL}. Since the 3D scanner simultaneously acquires a 2D scan, there is direct correspondence between their landmark points $p_{i} \in \mathbb{R}^{2}$ and $P_{i} \in \mathbb{R}^{3}$ respectively. So, the (x, y) co-ordinates for $p_{i}$, detected using Dlib \cite{Dlib}, are used to retrieve the (X, Y, Z) co-ordinates for $P_{i}$. For any scanner mis-registration (no valid $P_{i}$ for a $p_{i}$) we interpolate $P_{i}$ using its valid neighboring points. Since $P_{i}$ belongs to the same 3D plane as its immediate neighbors, we compute its (X, Y) using $\Delta \text{x} \rightarrow \Delta \text{X}$ and $\Delta \text{y} \rightarrow \Delta \text{Y}$ correspondences and solve the plane equation with (X, Y) to retrieve its Z co-ordinate.

We create a mesh of the face mask by triangulating the landmarks $P_{i}$ of the 3D model (point cloud). This mesh (107 triangles) is further refined by calculating the centroid of each triangle and re-triangulating with the new points. This re-triangulation step is performed twice to generate a denser mesh of the same facial mask (973 triangles). Using the same $p_{i} \rightarrow P_{i}$ correspondences, we map the synthetic texture $\text{{\bf I}}_{s}$ on this mesh and render the synthetic face mask $\text{{\bf M}}_{s}$ in 3D. As we do not possess the correspondence between the forehead and background points of a face scan and 3D model, $\text{{\bf M}}_{s}$ is generated with a black background. For each 3D model, we render $\text{{\bf M}}_{s}$ with $\theta$ = [0\degree, $\pm$ 30\degree, $\pm$ 60\degree,$\pm$ 90\degree]. Since we map the same texture on different 3D models, the overall appearance of the face masks remains the same with variations in its shape. So we assign to $\text{{\bf M}}_{s}$ the same label as $\text{{\bf I}}_{s}$ and generate 21 (7 poses$\times$3 models) new views of the synthetic identity. A final set of 2,060,992 face masks is obtained for the 12,338 synthetic identities via this rendering scheme (Table \ref{Tab:SREFIData}). Sample 2D face images of such synthetic identities, with their 3D renderings, can be seen in Figures \ref{fig:SuppFA}, \ref{fig:SuppFW}, \ref{fig:SuppMA}, and \ref{fig:SuppMW} respectively. The complete dataset, along with the 3D head models, can be accessed by clicking on the \emph{Notre Dame Synthetic Face Dataset} link on this page: \url{https://cvrl.nd.edu/projects/data/}. 

\section{Experiments \& Results}\label{ExpFull}
We perform two quantitative benchmark experiments to evaluate the effectiveness of our synthetic face images as training data supplement for CNNs and as impostors in face verification, and a qualitative comparison with three popular GAN based models.\\

\begin{table*}
\begin{center}
\captionsetup{justification=centering}
\caption{Distribution and Performance of Datasets in Experiment 1}
\begin{small}
\begin{tabular}{  | c | c| c| c|  }
\hline
{\bf Training Data} & \begin{tabular}[x]{@{}c@{}}{\bf CW \cite{CASIA} Images}\\(identities)\end{tabular} & \begin{tabular}[x]{@{}c@{}}{\bf Synthetic images}\\(identities)\end{tabular} & \begin{tabular}[x]{@{}c@{}}{\bf IJB-B \cite{IJBB} Performance}\\(TAR@FAR = 0.01)\end{tabular} \\
 %Ethnicity & Male images & Female images \\
\hline
\hline
  Dataset 1  &  494,414 (10,575)  &  0  &  0.891\\
  \hline
  Dataset 2  &  329,609 (7,050)  &  164,807 (3,525)  &  0.895\\
  \hline 
  Dataset 3  &  164,807 (3,525)  &  329,625 (7,050)  &  0.890\\
  \hline 
  Dataset 4  &  329,609 (7,050)  &  329,634 (3,525)  &  0.902\\
  \hline 
  Dataset 5  &  329,609 (7,050)  &  329,625 (7,050)  &  0.909\\
  \hline 
  {\bf Dataset 6}  &  {\bf 494,414 (10,575)}  &  {\bf 494,414 (10,575)}  &  {\bf 0.917}\\
  \hline 
  Dataset 7  &  0  &  494,414 (10,575)  &  0.774\\
%   \hline
%   Total  &  9011 (778)  &  6796 (674) \\
\hline
\end{tabular}
\label{Tab:Exp1}
\end{small}
\end{center}
\end{table*}

\noindent {\bf Experiment 1: Supplementing Training Data.} In this experiment, we aim to answer the following questions:\\
(1) Can our synthetic face images be used to supplement an existing face dataset for CNN training? (similar to \cite{SREFI1,MasiAug}).\\
(2) Is a synthetic face image nearly as effective (leads to the same level of accuracy) in training a CNN compared to a real face image?\\
(3) Can a CNN be trained effectively on only a set of synthetic face images?

To answer these questions, we prepare seven different training datasets using real face images (masked with black background) from the CASIA-WebFace (CW) dataset \cite{CASIA} and randomly drawn face images from our synthetic 3D dataset (Table \ref{Tab:SREFIData}). We mask the context (forehead, hair, neck, etc.) and background pixels in the face images from CW to maintain consistency with the synthetic images generated by our method. The distribution in the datasets can be seen in Table \ref{Tab:Exp1}. We fine-tune the VGG-16 network \cite{Simonyan14c} with these seven datasets in seven separate training sessions using Caffe \cite{Caffe}. For each dataset in Table \ref{Tab:Exp1}, 90\% of the data is used for training and the rest for validation, with each image resized to 224$\times$224 prior to training. We use SGD \cite{bottou2010large} for training each network with the same batch size = 64, base learning rate = 0.01, gamma = 0.1 and a step size of 50K training iterations. We stop training a network when its validation loss plateaus across a training epoch. 

\begin{figure*}
\centering
   \includegraphics[width=1.0\linewidth]{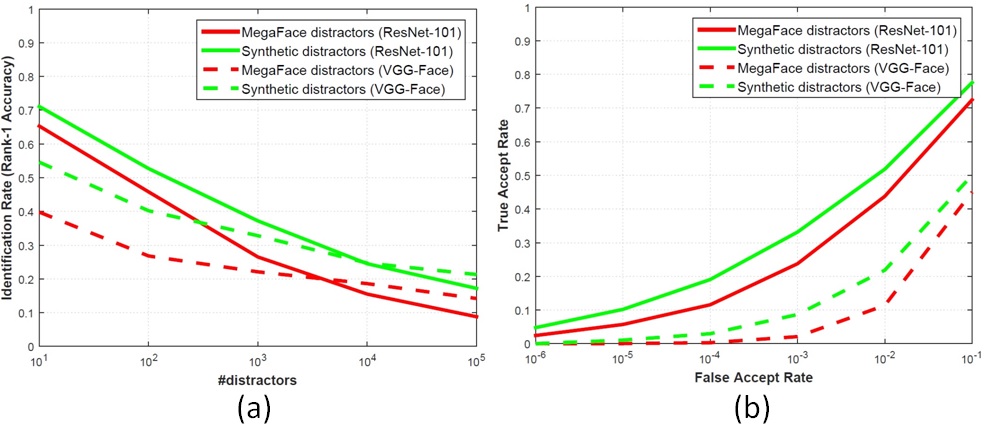}
   \caption{Experiment 2 results on the FG-Net dataset \cite{FGNet} - (a) Identification accuracy with variable real and synthetic distractor gallery size, (b) verification performance with real and synthetic 10K distractors, using ResNet-101 \cite{ResNet,MasiAug} and VGG-Face \cite{VGG} networks.}
\label{fig:Distractor}
%\label{fig:onecol}
\end{figure*}

For testing each trained network, we use the IJB-B verification protocol \cite{IJBB} as our performance metric. Each still image or video frame from a \emph{template} is first aligned about its eye center, with the face region masked out using its landmarks \cite{Dlib}. We feed the masked images to each of the seven trained networks and extract its \emph{fc7} layer descriptor. We generate an average feature vector for a template using \emph{video} and \emph{media pooling} operations, described in \cite{masiFG17}, and match a pair of templates using a simple correlation co-efficient metric between their feature vectors. The ROC performance of each network can be found in Table \ref{Tab:Exp1}. Results show that supplementing an existing dataset with our synthetic face images invariably improves CNN performance (Dataset 4, 5 and 6). However, a synthetic image does not appear to have the same value as a real image (Dataset 7). This can be attributed to the uniformity in lighting and expression in our synthetic face images while face images of both CW \cite{CASIA} and IJB-B \cite{IJBB} have plenty of variation in these areas. Another interesting observation is that the network trained with Dataset 5 (14,100 identities) slightly outperforms the network trained with Dataset 4 (10,575 identities) although they have the same number of images. This suggests wider training datasets \emph{i.e.} more identities, are indeed beneficial to CNN performance \cite{DosDonts}.\\

\noindent {\bf Experiment 2: Distractor Set.} In this experiment, we seek to answer these questions:\\
(1) Can our synthetic face images be used as distractors to influence the recognition accuracy of a trained CNN? (similar to \cite{MegaFace,MF2}).\\
(2) How does the size of this distractor set affect CNN performance?\\
(3) Is a synthetic distractor set as effective as a real distractor set (containing real face images, like MegaFace \cite{MegaFace}) of the same size?

For our experiments, the FG-Net dataset \cite{FGNet}, containing 982 images of 82 identities, is used as the probe set while we introduce two kinds of distractor galleries: $\text{{\bf D}}_{S}$ containing synthetic face images from Table \ref{Tab:SREFIData} and $\text{{\bf D}}_{M}$ with randomly drawn real face images from the MegaFace dataset \cite{MegaFace}. For a given probe identity with \emph{M} photos, we add each photo to the gallery and use each of the other (\emph{M} - 1) photos as probes (similar to the MegaFace protocol \cite{MegaFace}). We repeat this process for all the \emph{N} identities in the probe set.  

We use \emph{fc7} layer descriptor from pre-trained VGG-Face \cite{VGG} and \emph{pool5} layer descriptor from ResNet-101 \cite{ResNet}, pre-trained on CASIA-WebFace \cite{CASIA} following the augmentation described in \cite{MasiAug}, as feature vectors for all face images. It is to be noted that we do not repose the test images, as done in \cite{MasiAug}, when using ResNet-101. The metric for comparison is $L_{2}$ distance, similar to \cite{MegaFace}. We perform two different experiments: (1) a rank-based identification experiment (CMC) with $\text{{\bf D}}_{S}$ and $\text{{\bf D}}_{M}$ containing [$10^{1}$, $10^{2}$, $10^{3}$, $10^{4}$, $10^{5}$] distractors, (2) a verification experiment (ROC) with $\text{{\bf D}}_{S}$ and $\text{{\bf D}}_{M}$ containing $10^{4}$ distractors. The results are shown in Figure \ref{fig:Distractor}:a and \ref{fig:Distractor}:b respectively. 
% We crop the the face images from FG-Net and  as it is trained with face images at different poses overlaid on a black background, a similar setting to ours.

As depicted in the figure, our synthetic face images can be effectively used as distractors in face recognition experiments. But they are not 1:1 effective as the MegaFace images in their role as distractors, as evident from the gap between the green and red curves. This can be attributed to the uniform black background and neutral expression of our face images compared to the wide variation in the MegaFace \cite{MegaFace} images. However, since our face synthesis process demands less resources compared to downloading data (face images) from the web or acquiring new face images in a collection effort, we can always increase the synthetic distractor set size by adding in more 3D models or gallery base faces. \\

\begin{table}
\begin{center}
\captionsetup{justification=centering}
\caption{Comparison with GAN models based on training and synthesis time.}
\begin{small}
\begin{tabular}{  | c | c| c| c|  }
\hline
\begin{tabular}[x]{@{}c@{}}{\bf Synthesis}\\{\bf Method}\end{tabular} & {\bf GPU} & \begin{tabular}[x]{@{}c@{}}{\bf Training}\\ {\bf Time}\\(hours)\end{tabular} & \begin{tabular}[x]{@{}c@{}}{\bf Generation}\\{\bf Time}\\(seconds)\end{tabular} \\
 %Ethnicity & Male images & Female images \\
\hline
\hline
  DCGAN \cite{DCGAN}  & Yes &  9.8  &  0.58 \\
  \hline
  BEGAN \cite{BEGAN}  & Yes & 11.6  &  0.47 \\
  \hline 
  PGGAN \cite{ProgressiveGAN} & Yes &  60.9  &  0.35 \\
  \hline 
\hline
  Ours (2D texture) & No &  0.004  &  1.36 \\
  \hline
  Ours (3D mask) & No &  0  &  2.85 \\
%   \hline
%   Total  &  9011 (778)  &  6796 (674) \\
\hline
\end{tabular}
\label{Tab:Exp3}
\end{small}
\end{center}
\vspace{-1cm}
\end{table}

% \begin{figure}
% \begin{floatrow}
% \ffigbox{%
%   \includegraphics[width=1.0\linewidth]{Images/GAN_Comparison_2.jpg}%
% }{%
%   \caption{Comparison with GAN models based on visual quality.}%
% }
% \label{fig:GAN_Comp}
% \capbtabbox{%
% \begin{tiny}
% \begin{tabular}{  | c| c| c| c|  }
% \hline
% \begin{tabular}[x]{@{}c@{}}{\bf Synthesis}\\{\bf Method}\end{tabular} & {\bf GPU} & \begin{tabular}[x]{@{}c@{}}{\bf Training Time}\\(hours)\end{tabular} & \begin{tabular}[x]{@{}c@{}}{\bf Generation Time}\\(seconds)\end{tabular} \\
%  %Ethnicity & Male images & Female images \\
% \hline
% \hline
%   DCGAN \cite{DCGAN}  & Yes &  9.8  &  0.58 \\
%   \hline
%   BEGAN \cite{BEGAN}  & Yes & 11.6  &  0.47 \\
%   \hline 
%   PGGAN \cite{ProgressiveGAN} & Yes &  60.9  &  0.35 \\
%   \hline 
% \hline
%   Ours (2D texture) & No &  0  &  1.36 \\
%   \hline
%   Ours (3D mask) & No &  0  &  2.85 \\
% %   \hline
% %   Total  &  9011 (778)  &  6796 (674) \\
% \hline
% \end{tabular}
% \end{tiny}
% }{%
%   \caption{Comparison with GAN models based on training and synthesis time.}%
% }
% \label{Tab:Exp3}
% \end{floatrow}
% \end{figure}

\begin{figure*}
\centering
   \includegraphics[width=1.0\linewidth]{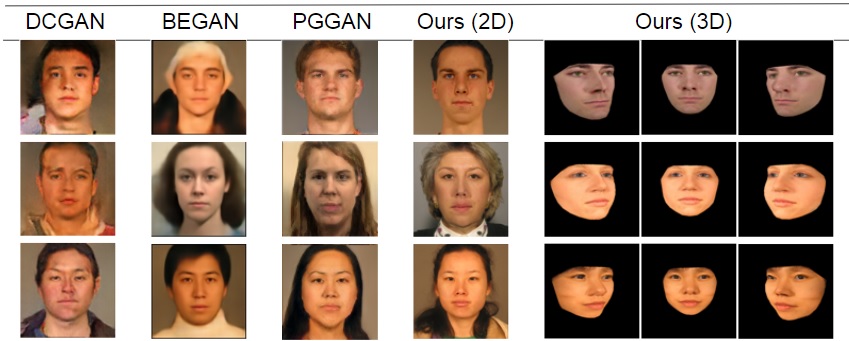}
   \caption{Comparison with GAN models based on visual quality.}
\label{fig:GAN_Comp}
%\label{fig:onecol}
\end{figure*}

\noindent {\bf Experiment 3: Comparison with GANs.} To compare with our method, we choose three popular GAN models for face image synthesis - DCGAN \cite{DCGAN}, BEGAN \cite{BEGAN} and the recently released PGGAN \cite{ProgressiveGAN}. To level the playing field resolution wise, we resize the 15,807 real face images from Table \ref{Tab:SREFIData} to 128$\times$128 and train the Tensorflow implementation of each GAN model with them for 200 epochs using a single NVIDIA Titan Xp GPU. For the PGGAN \cite{ProgressiveGAN} training session we used a lower batch size (16) compared to DCGAN \cite{DCGAN} and BEGAN \cite{BEGAN} (64) to accommodate for its much higher memory requirements. Each trained model is then used to generate synthetic face images, 128$\times$128 in size. We also generate 128$\times$128 synthetic face images, using our pipeline, from the same 15,807 face images (used as gallery). We compare the training and synthesis time required by all four methods (Table \ref{Tab:Exp3}), along with a visual comparison of the corresponding synthesis results (Figure \ref{fig:GAN_Comp}).

As expected the GAN models take a considerable amount of training time, especially the PGGAN framework \cite{ProgressiveGAN} due to its multi-phase training regime. Our method, in comparison, only uses pre-trained network models and therefore requires no GPU resources. The only training component of our method is to train the linear SVM for quality estimation of the synthetic 2D textures, however that requires only a few seconds. The most expensive steps of our pipeline are blending and 3D triangulation, which can vary with the target resolution. A disadvantage of our current method is that it cannot render multi-pose face images with forehead and background in 3D due to lack of 2D $\rightarrow$ 3D correspondence between the 2D synthetic textures (with background) and 3D models. We plan to address that as a future work.

\section{Conclusion}
In this paper, we propose a method to generate synthetic face images of real and synthetic identities with different shape, pose and resolution. Our method is training-free and requires little data and time compared to GANs \cite{GAN} to generate high quality face images. We formulate a set of quantitative benchmark experiments that can be potentially used to assess quality metrics of synthetic face images generated using different algorithms. We generate a dataset containing over 2M face images of 12K synthetic identities which can be used as supplemental data while training a CNN, or as a distractor set when performing verification experiments with CNNs. The dataset containing 2M synthetic face images and 8K 3D head models has been released, and can be downloaded by accessing the \emph{Notre Dame Synthetic Face Dataset} link from this webpage: \url{https://cvrl.nd.edu/projects/data/}.

A possible extension of this work can be to hallucinate realistic context (forehead, hair, neck, etc.) and background pixels on to our synthetic face masks, learned at multiple resolutions using GANs \cite{GAN}, and compare how the presence of background pixels in the training data affects CNN performance instead of just using face masks. We also plan to quantitatively analyze the influence of rendering the same 2D face texture (both real and synthetic) with different 3D head models on face verification. Finally, we would like to perform quantitative experiments, similar to experiments 1 and 2 in this paper, with synthetic face images generated using different GAN models \cite{DCGAN,BEGAN,ProgressiveGAN} and compare them with our method.

\noindent {\bf Acknowledgements}: The authors would like to thank Dr. Dodi Heryadi, Dr. Scott Hampton, and Dr. Paul Brenner, associated with the Center for Research Computing (CRC) at Notre Dame, for their help in rapid generation of face images on the grid. We also thank Dr. Chaoli Wang, Sreya Banerjee and Aparna Bharati for their helpful suggestions, and the NVIDIA Corporation for generously providing hardware support. 
% The contributions of this paper are as follows:\\
% 1) We propose a method to generate synthetic face images of real and synthetic identities with different shape, pose and resolution. Our method requires very little training data and execution time compared to GANs \cite{GAN} to generate high quality face images.\\
% 2) We formulate a set of quantitative benchmark experiments that can be potentially used to assess quality metrics of synthetic face images generated using different algorithms.\\
% 3) We generate a dataset containing over 2M face images of 12K synthetic identities which can be used as supplemental data while training a CNN, or as a distractor set when performing verification experiments with CNNs. We plan to release the dataset with this paper.\\
% 4) We plan to release the set of 3D head models used for 3D rendering with this paper.

{\small
\bibliographystyle{ieee}
\bibliography{egbib}

\begin{thebibliography}{10}\itemsep=-1pt

\bibitem{OpenGL}
Opengl.
\newblock In {\em \url{https://www.opengl.org/}}.

\bibitem{AbdAlmageed2016multipose}
W.~AbdAlmageed, Y.~Wu, S.~Rawls, S.~Harel, T.~Hassner, I.~Masi, J.~Choi,
  J.~Lekust, J.~Kim, P.~Natarajan, R.~Nevatia, and G.~Medioni.
\newblock Face recognition using deep multi-pose representations.
\newblock In {\em WACV}, 2016.

\bibitem{FaceAge}
G.~Antipov, M.~Baccouche, and J.~L. Dugelay.
\newblock Face aging with conditional generative adversarial networks.
\newblock {\em ICIP}, 2017.

\bibitem{SREFI1}
S.~Banerjee, J.~Bernhard, W.~Scheirer, K.~Bowyer, and P.~Flynn.
\newblock Srefi: Synthesis of realistic example face images.
\newblock In {\em IJCB}, 2017.

\bibitem{Vito}
S.~Banerjee, J.~Brogan, J.~Krizaj, A.~Bharati, B.~RichardWebster, V.~Struc,
  P.~Flynn, and W.~Scheirer.
\newblock To frontalize or not to frontalize? do we really need elaborate
  pre-processing to improve face recognition?
\newblock {\em WACV}, 2018.

\bibitem{DosDonts}
A.~Bansal, C.~Castillo, R.~Ranjan, and R.~Chellappa.
\newblock The do's and don'ts for cnn-based face verification.
\newblock {\em ICCV Workshops}, 2017.

\bibitem{UMDFaces}
A.~Bansal, A.~Nanduri, C.~D. Castillo, R.~Ranjan, and R.~Chellappa.
\newblock Umdfaces: An annotated face dataset for training deep networks.
\newblock {\em IJCB}, 2017.

\bibitem{PixelNN}
A.~Bansal, Y.~Sheikh, and D.~Ramanan.
\newblock Pixelnn: Example-based image synthesis.
\newblock {\em ICLR}, 2018.

\bibitem{CVAEGAN}
J.~Bao, D.~Chen, F.~Wen, H.~Li, and G.~Hua.
\newblock Cvae-gan: Fine-grained image generation through asymmetric training.
\newblock {\em ICCV}, 2017.

\bibitem{BEGAN}
D.~Berthelot, T.~Schumm, and L.~Metz.
\newblock Began: Boundary equilibrium generative adversarial networks.
\newblock {\em arXiv:1703.10717}.

\bibitem{Bitouk08}
D.~Bitouk, N.~Kumar, S.~Dhillon, S.~Belhumeur, and S.~K. Nayar.
\newblock Face swapping: Automatically replacing faces in photographs.
\newblock {\em SIGGRAPH}, 2005.

\bibitem{bottou2010large}
L.~Bottou.
\newblock Large-scale machine learning with stochastic gradient descent.
\newblock In {\em COMPSTAT}. 2010.

\bibitem{LapPyr}
P.~Burt and E.~Adelson.
\newblock The laplacian pyramid as a compact image code.
\newblock In {\em IEEE Trans. on Communications}, volume~31, pages 532--540,
  1983.

\bibitem{VGGFace2}
Q.~Cao, L.~Shen, W.~Xie, O.~M. Parkhi, and A.~Zisserman.
\newblock Vggface2: A dataset for recognizing faces across pose and age.
\newblock In {\em arXiv:1710.08092}.

\bibitem{Belanger}
F.~Cole, D.~Belanger, D.~Krishnan, A.~Sarna, I.~Mosseri, and W.~T. Freeman.
\newblock Face synthesis from facial identity features.
\newblock In {\em CVPR}, 2017.

\bibitem{SVM}
C.~Cortes and V.~Vapnik.
\newblock Support-vector networks.
\newblock {\em Machine learning}, 20(3):273--297, 1995.

\bibitem{TIST16}
C.~Ding and D.~Tao.
\newblock A comprehensive survey on pose-invariant face recognition.
\newblock {\em ACM Trans. on Intelligent Systems and Technology},
  7(3):37:1--37:42, 2016.

\bibitem{TMM15}
D.~Ding, C.and~Tao.
\newblock Robust face recognition via multimodal deep face representation.
\newblock {\em IEEE Trans. on Multimedia}, 17(11):2049--2058, 2015.

\bibitem{FaltemierCVIU}
T.~Faltemier, K.~Bowyer, and P.~Flynn.
\newblock Using multi-instance enrollment to improve performance of 3d face
  recognition.
\newblock {\em CVIU}, 112(2):114--125, 2008.

\bibitem{cGAN}
J.~Gauthier.
\newblock Conditional generative adversarial networks for convolutional face
  generation.
\newblock In {\em Tech Report}, 2015.

\bibitem{GAN}
I.~J. Goodfellow, J.~Pouget-Abadie, M.~Mirza, B.~Xu, D.~Warde-Farley, S.~Ozair,
  A.~C. Courville, and Y.~Bengio.
\newblock Generative adversatial nets.
\newblock In {\em NIPS}, 2014.

\bibitem{Affact}
M.~Gunther, A.~Rozsa, and T.~Boult.
\newblock Affact: Alignment free facial attribute classification technique.
\newblock In {\em IJCB}, 2017.

\bibitem{FaceSketch}
Q.~Guo, C.~Zhu, Z.~Xia, Z.~Wang, and Y.~Liu.
\newblock Attribute-controlled face photo synthesis from simple line drawing.
\newblock {\em arXiv:1702.02805}, 2017.

\bibitem{HassFront}
T.~Hassner, S.~Harel, E.~Paz, and R.~Enbar.
\newblock Effective face frontalization in unconstrained images.
\newblock In {\em CVPR}, 2015.

\bibitem{PACNN}
K.~He and X.~Xue.
\newblock Facial landmark localization by part aware deep convolutional
  network.
\newblock In {\em Pacific Rim Conference on Multimedia}, 2016.

\bibitem{ResNet}
K.~He, X.~Zhang, S.~Ren, and J.~Sun.
\newblock Deep residual learning for image recognition.
\newblock {\em CVPR}, 2016.

\bibitem{SketchBMVC17}
C.~Hu, D.~Li, Y.~Z. Song, and T.~M. Hospedales.
\newblock Now you see me: Deep face hallucination for unviewed sketches.
\newblock {\em BMVC}, 2017.

\bibitem{SENet}
J.~Hu, L.~Shen, and G.~Sun.
\newblock Squeeze-and-excitation networks.
\newblock {\em arXiv:1709.01507}, 2017.

\bibitem{LFW}
G.~B. Huang, M.~Ramesh, T.~Berg, and E.~Learned-Miller.
\newblock Labeled faces in the wild: A database for studying face recognition
  in unconstrained environments.
\newblock In {\em Tech Report 07--49}, 2007.

\bibitem{HeFrontal}
R.~Huang, S.~Zhang, T.~Li, and R.~He.
\newblock Beyond face rotation: Global and local perception gan for
  photorealistic and identity preserving frontal view synthesis.
\newblock {\em ICCV}, 2017.

\bibitem{Bulat3D}
A.~S. Jackson, A.~Bulat, V.~Argyriou, and G.~Tzimiropoulos.
\newblock Large pose 3d face reconstruction from a single image via direct
  volumetric cnn regression.
\newblock {\em ICCV}, 2017.

\bibitem{Caffe}
Y.~Jia, E.~Shelhamer, J.~Donahue, S.~Karayev, J.~Long, R.~Girshick,
  S.~Guadarrama, and T.~Darrell.
\newblock Caffe: Convolutional architecture for fast feature embedding.
\newblock In {\em ACM MM}, 2014.

\bibitem{ProgressiveGAN}
T.~Karras, T.~Aila, S.~Laine, and J.~Lehtinen.
\newblock Progressive growing of gans for improved quality, stability, and
  variation.
\newblock {\em ICLR}, 2018.

\bibitem{IraPortrait}
I.~Kemelmacher-Shlizerman.
\newblock Transfiguring portraits.
\newblock {\em SIGGRAPH}, 2016.

\bibitem{MegaFace}
I.~Kemelmacher-Shlizerman, S.~Seitz, D.~Miller, and E.~Brossard.
\newblock The megaface benchmark: 1 million faces for recognition at scale.
\newblock In {\em CVPR}, 2016.

\bibitem{Dlib}
D.~E. King.
\newblock Dlib-ml: A machine learning toolkit.
\newblock In {\em Journal of Machine Learning Research}, volume~10, pages
  1755--1758, 2009.

\bibitem{IJBA}
B.~F. Klare, B.~Klein, E.~Taborsky, A.~Blanton, J.~Cheney, K.~Allen,
  P.~Grother, A.~Mah, and A.~K. Jain.
\newblock Pushing the frontiers of unconstrained face detection and
  recognition: Iarpa janus benchmark a.
\newblock In {\em CVPR}, 2015.

\bibitem{DLNature}
Y.~LeCun, Y.~Bengio, and G.~Hinton.
\newblock Deep learning.
\newblock {\em Nature}, 521(7553):436--444, 2015.

\bibitem{coGAN}
M.-Y. Liu and O.~Tuzel.
\newblock Coupled generative adversarial networks.
\newblock In {\em arXiv:1606.07536}.

\bibitem{ICME05}
W.~Liu, D.~Lin, and X.~Tang.
\newblock Neighbor combination and transformation for hallucinating faces.
\newblock In {\em ICME}, 2005.

\bibitem{SynthesisSurvey}
Z.~Lu, Z.~Li, J.~Cao, R.~He, and Z.~Sun.
\newblock Recent progress of face image synthesis.
\newblock {\em arXiv:1706.04717}.

\bibitem{masiFG17}
I.~Masi, T.~Hassner, A.~T. Tran, and G.~Medioni.
\newblock Rapid synthesis of massive face sets for improved face recognition.
\newblock {\em FG}, 2017.

\bibitem{masi2016cvpr}
I.~Masi, S.~Rawls, G.~Medioni, and P.~Natarajan.
\newblock Pose-{A}ware {F}ace {R}ecognition in the {W}ild.
\newblock In {\em CVPR}, 2016.

\bibitem{MasiAug}
I.~Masi, A.~T. Tran, J.~T. Leksut, T.~Hassner, and G.~Medioni.
\newblock Do we really need to collect millions of faces for effective face
  recognition?
\newblock In {\em ECCV}, 2016.

\bibitem{SIGGRAPH09}
U.~Mohammed, S.~J.~D. Prince, and J.~Kautz.
\newblock Visio-lization: Generating novel facial images.
\newblock {\em SIGGRAPH}, 2009.

\bibitem{ACCV14}
S.~Mosaddegh, L.~Simon, and F.~Jurie.
\newblock Photorealistic face de-identification by aggregating donors' face
  components.
\newblock In {\em ACCV}, 2014.

\bibitem{MF2}
A.~Nech and I.~Kemelmacher-Shlizerman.
\newblock Level playing field for million scale face recognition.
\newblock {\em CVPR}, 2017.

\bibitem{FaceSwap}
Y.~Nirkin, I.~Masi, A.~T. Tran, T.~Hassner, and G.~Medioni.
\newblock On face segmentation, face swapping, and face perception.
\newblock {\em arXiv:1704.06729}, 2017.

\bibitem{FGNet}
G.~Panis, A.~Lanitis, N.~Tsapatsoulis, and T.~Cootes.
\newblock Overview of research on facial ageing using the fg-net ageing
  database.
\newblock {\em IET Biometrics}, 5(2):37--46, 2016.

\bibitem{VGG}
O.~M. Parkhi, A.~Vedaldi, and A.~Zisserman.
\newblock Deep face recognition.
\newblock In {\em BMVC}, 2015.

\bibitem{SREFIDonor}
P.~Phillips, P. J.and~Flynn and K.~Bowyer.
\newblock Lessons from collecting a million biometric samples.
\newblock {\em Image and Vision Computing}, 2016.

\bibitem{DCGAN}
A.~Radford, L.~Metz, and S.~Chintala.
\newblock Unsupervised representation learning with deep convolutional
  generative adversarial networks.
\newblock In {\em ICLR}, 2016.

\bibitem{PaSC}
W.~Scheirer, P.~Flynn, C.~Ding, G.~Guo, V.~Struc, M.~Al~Jazaery, K.~Grm,
  S.~Dobrisek, D.~Tao, Y.~Zhu, J.~Brogan, S.~Banerjee, A.~Bharati, and
  B.~RichardWebster.
\newblock Report on the btas 2016 video person recognition evaluation.
\newblock In {\em BTAS}, 2016.

\bibitem{Google_FaceNet}
F.~Schroff, D.~Kalenichenko, and J.~Philbin.
\newblock Facenet: A unified embedding for face recognition and clustering.
\newblock In {\em CVPR}, 2015.

\bibitem{Simonyan14c}
K.~Simonyan and A.~Zisserman.
\newblock Very deep convolutional networks for large-scale image recognition.
\newblock {\em ICLR}, 2015.

\bibitem{Facebook_Deepface}
Y.~Taigman, M.~Yang, M.~Ranzato, and L.~Wolf.
\newblock Deepface: Closing the gap to human-level performance in face
  verification.
\newblock In {\em CVPR}, 2014.

\bibitem{USC3DMM}
A.~T. Tran, T.~Hassner, I.~Masi, and G.~Medioni.
\newblock Regressing robust and discriminative {3D} morphable models with a
  very deep neural network.
\newblock {\em CVPR}, 2017.

\bibitem{DRGAN}
L.~Tran, X.~Yin, and X.~Liu.
\newblock Disentangled representation learning gan for pose-invariant face
  recognition.
\newblock In {\em CVPR}, 2017.

\bibitem{DFI}
P.~Upchurch, J.~Gardner, G.~Pleiss, R.~Pless, N.~Snavely, K.~Bala, and
  K.~Weinberger.
\newblock Deep feature interpolation for image content changes.
\newblock {\em CVPR}, 2017.

\bibitem{MAGAN}
R.~Wang, A.~Cully, H.~J. Chang, and Y.~Demiris.
\newblock Magan: Margin adaptation for generative adversarial networks.
\newblock {\em arXiv:1704.03817}, 2017.

\bibitem{IJBB}
C.~Whitelam, E.~Taborsky, A.~Blanton, B.~Maze, J.~Adams, T.~Miller, N.~Kalka,
  A.~K. Jain, J.~A. Duncan, K.~Allen, and et~al.
\newblock Iarpa janus benchmark-b face dataset.
\newblock In {\em CVPR Workshops}, 2017.

\bibitem{YoutubeFaces}
L.~Wolf, T.~Hassner, and I.~Maoz.
\newblock Face recognition in unconstrained videos with matched background
  similarity.
\newblock In {\em CVPR}, 2011.

\bibitem{LightCNN}
X.~Wu, R.~He, Z.~Sun, and T.~Tan.
\newblock A light cnn for deep face representation with noisy labels.
\newblock {\em arXiv:1511.02683}, 2017.

\bibitem{SIGGRAPH11}
F.~Yang, J.~Wang, E.~Shechtman, L.~Bourdev, and D.~Metaxas.
\newblock Expression flow for 3d-aware face component transfer.
\newblock {\em SIGGRAPH}, 2011.

\bibitem{CASIA}
D.~Yi, Z.~Lei, S.~Liao, and S.~Z. Li.
\newblock Learning face representation from scratch.
\newblock In {\em arXiv:1411.7923}.

\bibitem{LiuFrontal}
X.~Yin, X.~Yu, K.~Sohn, X.~Liu, and M.~Chandraker.
\newblock Towards large-pose face frontalization in the wild.
\newblock {\em ICCV}, 2017.

\bibitem{EBGAN}
J.~Zhao, M.~Mathieu, and Y.~LeCun.
\newblock Energy-based generative adversarial networks.
\newblock In {\em ICLR}, 2017.

\bibitem{NIPS17}
J.~Zhao, L.~Xiong, K.~Jayashree, J.~Li, F.~Zhao, Z.~Wang, S.~Pranata, S.~Shen,
  S.~Yan, and J.~Feng.
\newblock Dual-agent gans for photorealistic and identity preserving profile
  face synthesis.
\newblock In {\em NIPS}, 2017.

\end{thebibliography}
}

\end{document}